\documentclass[letterpaper]{article} 
\usepackage{aaai20_arxiv}  
\usepackage{times}  
\usepackage{helvet} 
\usepackage{courier}  
\usepackage[hyphens]{url}  
\usepackage{graphicx} 
\usepackage{graphicx}  
\usepackage{multirow}
\usepackage{amsmath}
\usepackage{amssymb}
\usepackage{subfigure}
\usepackage{algorithm}
\usepackage{algorithmicx}  
\usepackage[noend]{algpseudocode} 
\algrenewcommand\algorithmicrequire{\textbf{Input:}}
\algrenewcommand\algorithmicensure{\textbf{Output:}}
\newcommand{\etal}{\textit{et al}. }

\nocopyright
\title{Double Anchor R-CNN for Human Detection in a Crowd}
\author{\Large \textbf{
Kevin Zhang\textsuperscript{\rm 1}\thanks{Equal contribution},
Feng Xiong\textsuperscript{\rm 2}\footnotemark[1], Peize Sun\textsuperscript{\rm 3},
Li Hu\textsuperscript{\rm 4},
Boxun Li\textsuperscript{\rm 1},
Gang Yu\textsuperscript{\rm 1}}\\
\textsuperscript{\rm 1}Megvii Inc. (Face++)\quad
\textsuperscript{\rm 2}Tsinghua University\\
\textsuperscript{\rm 3}Xi'an Jiaotong University\quad
\textsuperscript{\rm 4}Zhejiang University\\
}

\begin{document}
\maketitle

\begin{abstract}

Detecting human in a crowd is a challenging problem due to the uncertainties of occlusion patterns. In this paper, we propose to handle the crowd occlusion problem in human detection by leveraging the head part. Double Anchor RPN is developed to capture body and head parts in pairs. A proposal crossover strategy is introduced to generate high-quality proposals for both parts as a training augmentation. Features of coupled proposals are then aggregated efficiently to exploit the inherent relationship. Finally, a Joint NMS module is developed for robust post-processing. The proposed framework, called Double Anchor R-CNN, is able to detect the body and head for each person simultaneously in crowded scenarios. State-of-the-art results are reported on challenging human detection datasets. Our model yields log-average miss rates (MR) of 51.79pp on CrowdHuman, 55.01pp on COCOPersons~(crowded sub-dataset) and 40.02pp on CrowdPose~(crowded sub-dataset), which outperforms previous baseline detectors by 3.57pp, 3.82pp, and 4.24pp, respectively. We hope our simple and effective approach will serve as a solid baseline and help ease future research in crowded human detection.

\end{abstract}

\section{Introduction}

Human detection serves as a key component for a wide range of real-world applications, such as advanced human-machine interactions, video surveillance or crowd analysis~\cite{Socially-Aware}. In recent years, the performance of human detectors has been rapidly improved with the development of deep convolutional neural networks (CNN)~\cite{Occlusion-aware,Bi-box,ImprovingOcclusion_PedDet}.

\begin{figure}[!t]
\centering
\includegraphics[width=8cm]{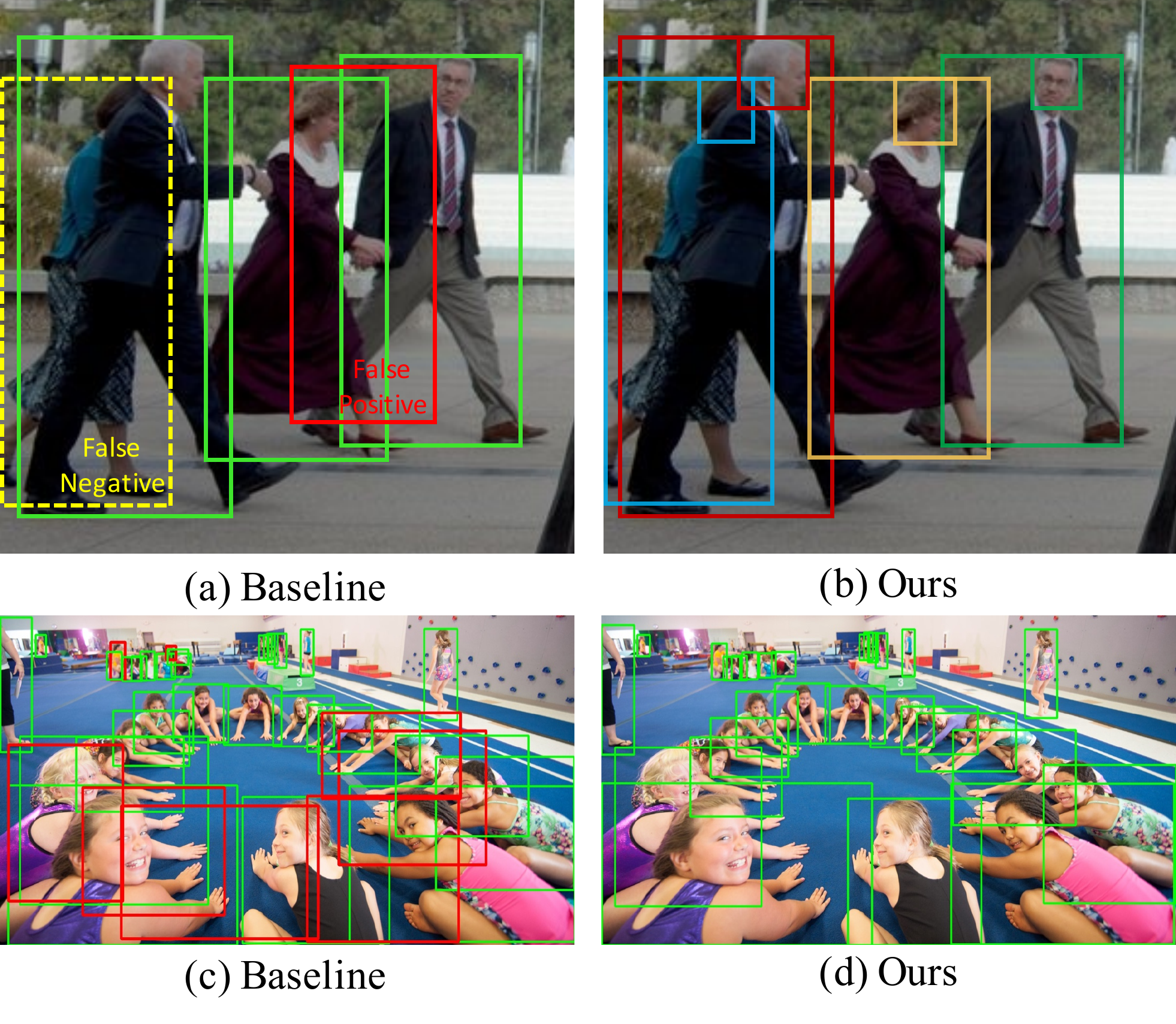}
\vspace{-0.5cm}
\caption{Visual comparisons of human detection results between the baseline and our Double Anchor R-CNN in crowded scenes. Green and red bounding boxes in subfigures (a, c and d) represent true positives and false positives for the human category, respectively (Only predictions with confidences above 0.5 are drawn). Pairs of bounding boxes with different colors stand for the results of our method in subfigure b. By leveraging the head part, our Double Anchor R-CNN is able to reduce false positives and recover heavily occluded humans in crowded scenarios.}
\vspace{-0.3cm}
\label{fig:fpn_fp}
\end{figure}

However, crowd occlusion~\cite{HowFar_PedDet} is a challenging problem for human detection systems. Examples are illustrated in Figure~\ref{fig:fpn_fp}. Crowded scenarios that happen frequently in real life bring several challenges for CNN-based detectors. First, there are large variations in scales, ratios, and poses in crowd scenes~\cite{HowFar_PedDet,Graininess-Aware} so robustness is a challenging issue.
Second, when people overlap largely with each other, semantic features of different instances also interweave and make detectors difficult to discriminate instance boundaries. As a result, detectors may treat the crowd as a whole, or shift the target bounding box to another person mistakenly~\cite{RepulsionLoss}. Finally, even though the detectors succeed to identify different human instances in a crowd, the highly overlapped results may also be suppressed by the post-processing of non-maximum suppression~(NMS). A higher NMS threshold is required to keep the crowded bounding boxes at the expense of bringing more false positives.

\begin{figure*}[!t]
\centering
\includegraphics[width=17cm]{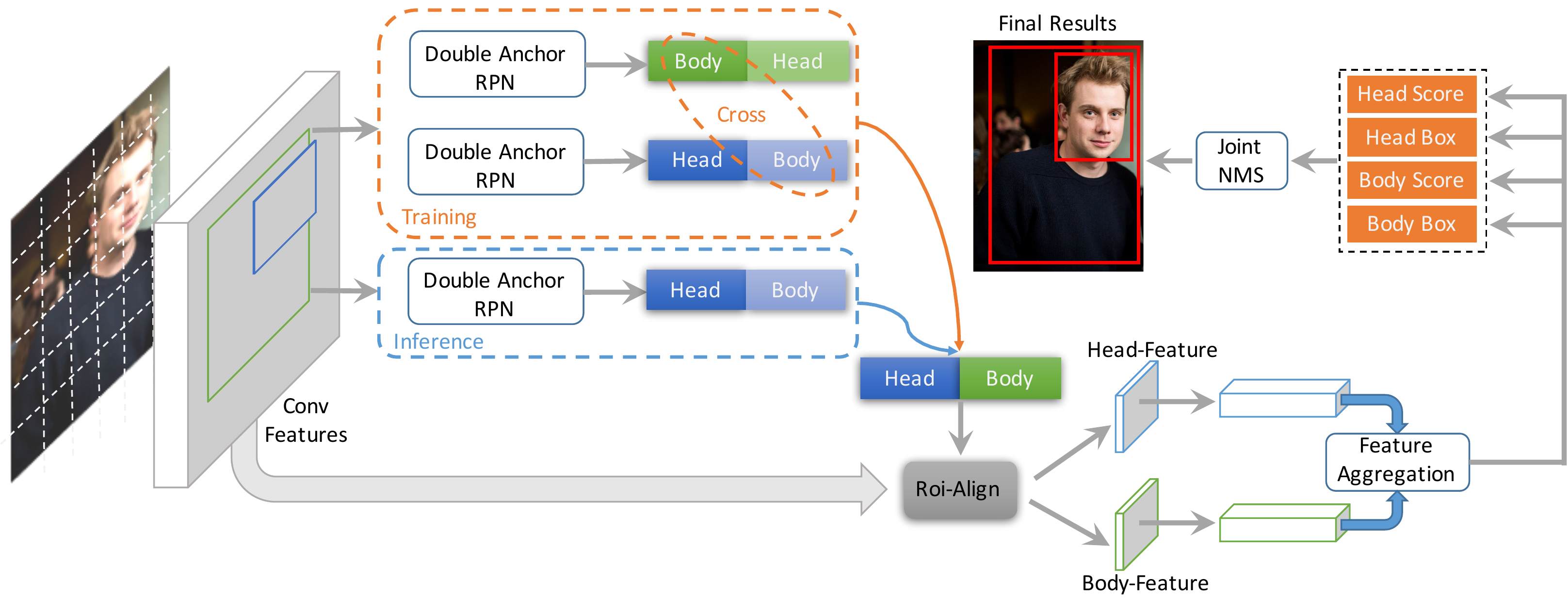}
\vspace{-0.5cm}
\caption{Illustration of the proposed framework. Head and body proposals are first extracted in pairs by Double Anchor RPN (described in Section~\ref{sec:RPN}). Both head-body and body-head branches are employed with different principal anchors. High-quality proposals will be crossed in the training phase for augmentation and then sent to the second stage with RoI Pool/RoI Align. Head and body features are extracted and aggregated before the final prediction. A Joint NMS module is deployed for final post-processing.}
\vspace{-0.3cm}
\label{fig:architecture}
\end{figure*}

A common solution to alleviate crowd occlusion problem is to focus on instance parts~\cite{Duan2010A,Mathias2014Handling,Ouyang2012A,Ouyang2014Joint,Tian2016Deep,Zhou2016Learning,Zhou2017Multi}. When a full-body detector fails to recognize an occluded person, the visible parts may give high confidences and guide the detector to discriminate instances crowded together. For part-based solutions, the reliability of part detectors is of great importance.
Most previous works~\cite{Duan2010A,Mathias2014Handling,Ouyang2012A,Ouyang2014Joint} generate part labels by leveraging the differences between visible-region and full-body bounding boxes for each person. These methods are usually designed for pedestrian detection, where most objects appear with similar poses and aspect ratios. However, we point out that the case is not suitable for human detection because of the large diversity of poses and occlusions, especially for visible-part human detection.

In this paper, we propose Double Anchor R-CNN to improve human detection in crowded scenes by detecting the body and head for each person at the same time. Compared with the human body, the head usually has a smaller scale, less overlap, and a better view in real-world images, and thus is more robust to pose variations and crowd occlusion. This is especially useful in crowded scenarios: Figure~\ref{fig:fpn_fp} shows a heavily crowded scene. The human detector is unable to discriminate instance boundaries since the parts of different instances interweave each other, which may lead to false positives. In this case, features of heads may significantly help discriminate different instances so that such false positive human detections that are not consistent with the head detections can be removed. Moreover, human detection is difficult in crowded situations due to the heavy occlusion or suppression by NMS, which may lead to false negatives. While the heads are still visible and overlap softly, which can notably help to recover heavily occluded humans.

The main contributions of this work are threefold:
\begin{itemize}
    \item We propose double anchor region proposal network (Double Anchor RPN) to detect human heads and bodies at the same time. The head and body of each person are naturally coupled and supply each other for human detection in a crowd. 
    \item A proposal crossover strategy is developed to generate high-quality proposals for both parts as a training augmentation. In addition, features of heads and bodies are aggregated efficiently to make the final prediction more reliable. A Joint NMS algorithm is introduced to suppress false positive results in a crowd and improve the robustness of the post-processing.
    \item State-of-the-art results are reported on various challenging human detection datasets. We have achieved a remarkable performance improvement of MR of at least 3pp on the CrowdHuman dataset, COCOPersons (crowded sub-dataset) and CrowdPose (crowded sub-dataset).
\end{itemize}

\section{Related Work}

\subsection{General Object Detection}
The advances in human detection systems have been driven by powerful baseline systems of general object detection. Modern object detection systems can be divided into two categories of one-stage detectors and two-stage detectors. Generally speaking, two-stage approaches on representative of Faster R-CNN~\cite{FasterRCNN} adopt a coarse-to-fine manner and focus on achieving top performances on various benchmarks~\cite{PASCAL,COCO}. As a comparison, one-stage approaches aim at achieving real-time speed while maintaining comparable performance~\cite{YOLO,YOLO9000,SSD}.

\subsection{Human Detection}
Besides detecting human as a simple category with general detectors, many works have been proposed to handle the occlusion and scale-variation problems in human detection~\cite{SAF-RCNN,Graininess-Aware,MSCNN,Bi-box,GuidedAttention}. SA-Fast R-CNN tries to handle the scale variation problem by extending Fast R-CNN with jointly training small-scale and large-scale networks~\cite{SAF-RCNN}. Lin~\etal propose an approach to incorporate fine-grained attention masks to extract better semantic features~\cite{Graininess-Aware}. Zhang~\etal propose an attention mechanism to focus on visible body regions instead of learning various parts~\cite{GuidedAttention}.

Several works have been proposed to detect human in a crowd by leveraging part-based detectors~\cite{Duan2010A,Enzweiler2010Multi,Wang2012A,Mathias2014Handling}. The part-based detectors assume that the visible parts are able to generate high confidence prediction and reveal the occluded body. 
Pioneer works usually train detectors of different parts independently. Later works exploit relationships between different parts by learning various part features in a joint way~\cite{Ouyang2014Joint,Tian2016Deep,Zhou2017Multi,Zhou2016Learning}. Most of the previous works generate part labels in a style of semi-supervised learning by comparing visible and full-body annotations of pedestrians~\cite{Tian2016Deep,Zhou2017Multi,Occlusion-aware}. However, the solution is hard to extend to human detection because of the huge diversity of poses and occlusions in real-world scenarios.

Special losses are also proposed to discriminate overlapped people in crowded scenes better. Wang~\etal propose repulsion loss to make surrounding proposals from different targets repel each other~\cite{RepulsionLoss}. 
Zhang~\etal design an aggregation loss to enforce proposals closer to the ground truth~\cite{Occlusion-aware}.
Besides, variants of NMS like Soft-NMS~\cite{SoftNMS} and Adaptive-NMS~\cite{AdaptiveNMS} are proposed to soften the 
sensitivity of NMS threshold in crowded scenarios.

\section{Double Anchor R-CNN}
The framework of Double Anchor R-CNN is illustrated in Figure~\ref{fig:architecture}.
The architecture is designed on top of the Feature Pyramid Network~(FPN)~\cite{FPN} and can be easily extended to other frameworks like Faster R-CNN and Mask R-CNN. Double Anchor R-CNN framework consists of the following phases: (i). a double anchor region proposal network to generate head and body proposals in pairs, (ii). a proposal crossover module to generate high-quality training samples for the R-CNN part, (iii). an aggregation module to fuse features of heads and bodies effectively, and (iv). a Joint NMS algorithm for post-processing. In this section, we introduce each part sequentially.

\begin{figure}[!t]
\centering
\includegraphics[width=8cm]{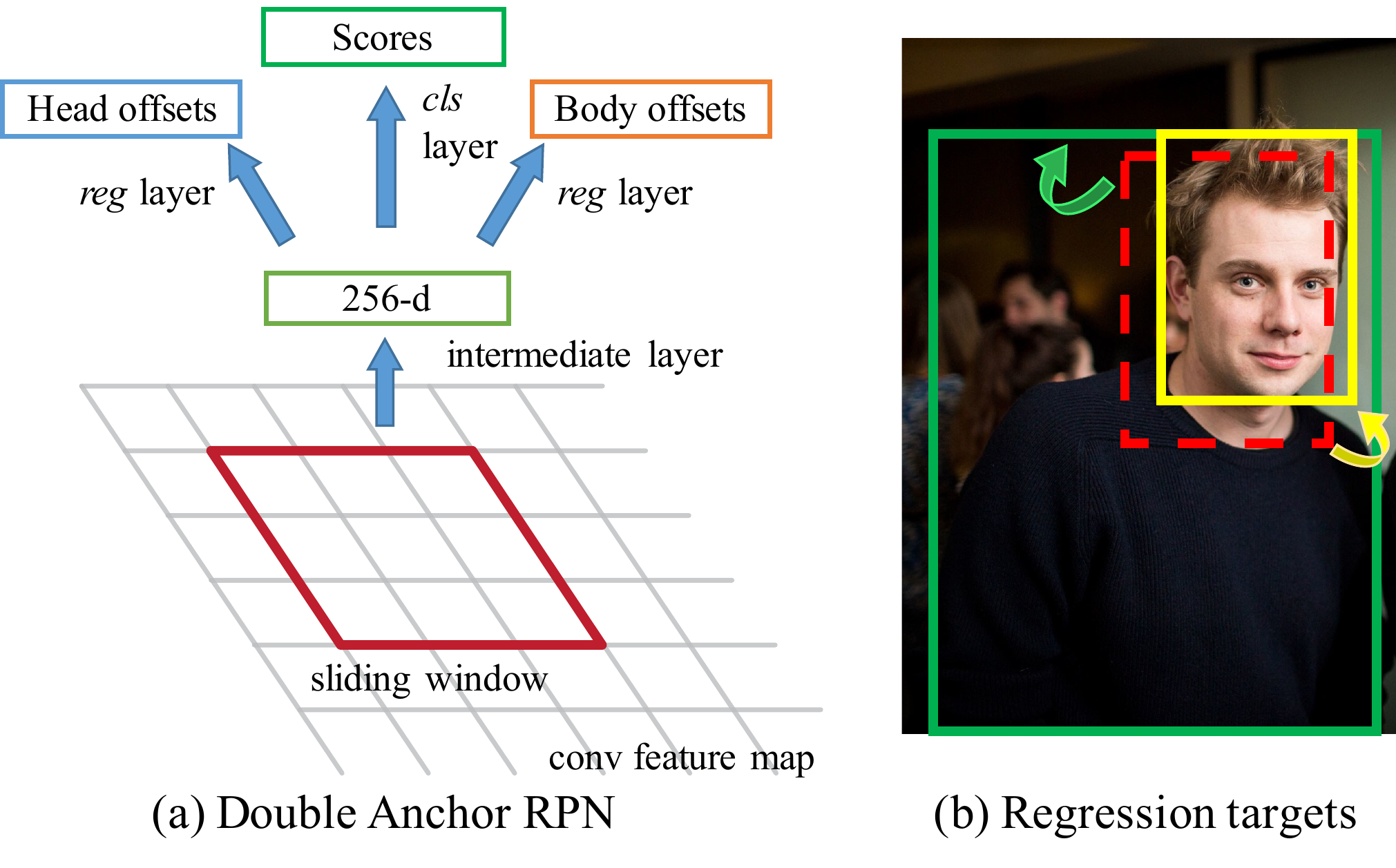}
\vspace{-0.3cm}
\caption{Illustration of Double Anchor RPN. (a). Compared with the basic region proposal network, an additional regression layer is added and the network predicts coupled offsets for both bodies and heads. (b). The red dashed bounding box stands for anchor. The network regresses offsets for head (yellow) and body (green) at the same time.}
\vspace{-0.3cm}
\label{fig:double_anchor}
\end{figure}

\subsection{Double Anchor RPN}
\label{sec:RPN}
The original region proposal network first slides a small network over the convolutional feature maps and regresses the target bounding boxes from pre-designed anchors. On top of that, Double Anchor RPN is conceptually simple: the network will regress both the head offsets and the body offsets for each human instance simultaneously from \textbf{the same anchor}. The method is shown in Figure~\ref{fig:double_anchor}. 

It should be noted that Double Anchor RPN requires to select one principal part in anchor matching. For example, we can set principal anchors to heads. Anchors overlap with the head ground-truths with high intersection-over-union (IoU) will be matched first. Then the network is forced to regress the attached body part based on the principal head anchors. We called this branch the head-body branch in this paper. To cover both parts better, two branches, i.e., the head-body branch and body-head branch, are employed in the framework. Each branch sets either heads or bodies as principal parts in Double Anchor RPN. Besides, Double Anchor RPN only predicts one classification score for each anchor, since region proposals are used to distinguish the foreground and background in a
class-agnostic style. Finally, the loss function for Double Anchor RPN module is designed as follows:
\begin{equation}
    L_{\rm{RPN}} = L_{\rm{cls}} + L^h_{\rm{reg}} + L^b_{\rm{reg}},
\end{equation}
where $L_{\rm{cls}}$ is the cross-entropy loss for classification of foreground and background. $L^h_{\rm{reg}}$ and $L^b_{\rm{reg}}$ are regression losses (e.g.~the Smooth $L_1$ loss) for head bounding boxes and body bounding boxes, respectively. 

For detailed implementation, we assign positive labels for anchors when the anchor overlaps with principal part ground-truth (e.g., the head ground-truth for the head-body branch) with an IoU larger than a threshold (0.7 in our work). Only one ground-truth with the highest IoU will be assigned as the target for offset regression. For positive anchors, we calculate the regression targets for both heads and bodies based on the same anchor.

\subsection{Proposal Crossover}
\label{sec:Crossover}
\begin{figure}[!t]
\centering
\includegraphics[width=8cm]{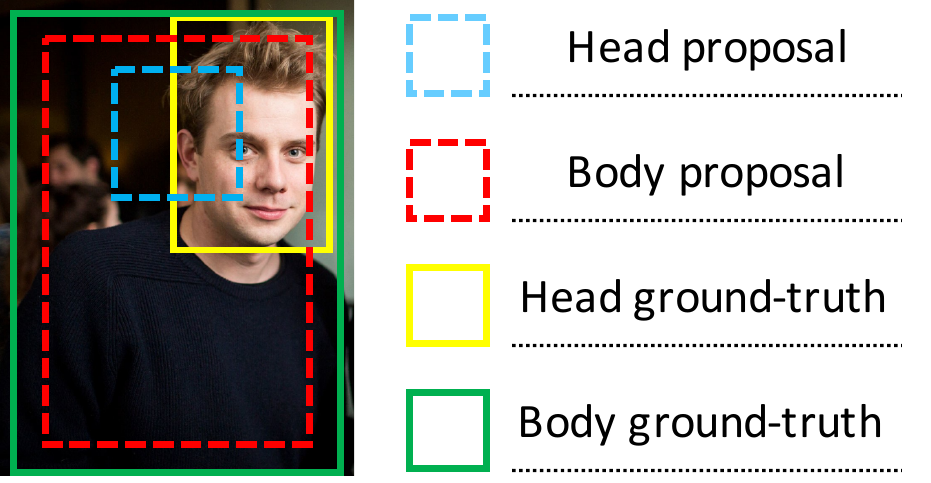}
\vspace{-0.3cm}
\caption{Illustration of the impact of proposal quality in RCNN stage. The body is set to be the principal part of Double Anchor RPN, making the body proposal enjoy a high overlap with the target. However, the noisy head proposal will lead to great difficult of bounding box regression.}
\vspace{-0.5cm}
\label{fig:proposal}
\end{figure}

Double Anchor RPN generates proposals in pairs of heads and bodies. The top confident pairs of proposals will be fed to the second RCNN stages with RoI module to predict final results. As mentioned in Cascade R-CNN~\cite{CascadeRCNN}, high-quality detection  \textbf{ requires sufficient high-quality positive samples}. However, as illustrated in Figure~\ref{fig:proposal}, we discover that the quality of the attached part is not guaranteed since Double Anchor RPN module only considers the principal part when assigning the pair label. 


A simple method to generate high-quality proposal pairs is to constrain the IoU thresholds for both parts in the pair. 
However, as discussed later in Section~\ref{sec:Selection}, this method does not work for Double Anchor R-CNN due to the insufficient positives that have qualified IoU for both parts. The network will be dominated by the noisy proposals and cannot discriminate ``good'' and ``bad'' proposals finally.


In order to generate more qualified proposal pairs, we introduce \textbf{a training augmentation strategy named Proposal Crossover}, which generates adequate augmented positive training samples by utilizing the complementary.
To be specific, we add a body-head branch as an augmentation along with the head-body branch as illustrated in Figure~\ref{fig:architecture}. First we can obtain the labels of the pairs from each branch, by calculating the overlaps between the principal parts of each branch and corresponding ground-truths. The pairs are regarded as positives if the overlap is larger than a threshold (0.5 in our work). It should be noted that the principal parts are qualified here but the attached parts are noisy since they are given the same positive labels without consideration of their own overlaps. Then we crossover the proposals between the head-body branch and a body-head branch to generate final paired proposals qualified for both parts. Overlaps between the attached part of head-body branch (a.k.a. body proposals) and the principal part of body-head branch (also the body proposals) are calculated. If the maximum overlap exceeds a certain threshold (0.5 in our work), the body proposals from the head-body branch will be replaced by the body proposals from the body-head branch with the maximum overlap. New pairs of proposals consist of original head proposals from the head-body branch, and crossover body proposals from the body-head branch are generated with good quality. Finally the crossover method generates adequate high-quality proposals for R-CNN and effectively leads to a better training procedure. 

It should be noted that the proposal crossover is not needed at inference time and will not introduce extra complexity since it only serves as an effective training augmentation for the R-CNN part.

\begin{algorithm}[!t]
    \caption{Joint NMS}
    \begin{algorithmic}[1]
    \Require
        \Statex $B^H=\{b^H_1,\cdots,b^H_N\}$: head boxes.
        \Statex $B^B=\{b^B_1,\cdots,b^B_N\}$: body boxes.
        \Statex $S^H=\{s^H_1,\cdots,s^H_N\}$: head scores.
        \Statex $S^B=\{s^B_1,\cdots,s^B_N\}$: body scores.
        \Statex $\Omega_H, \Omega_B$: NMS threshold for head and body.
        \Statex $\lambda$: weight of body scores.
    \Ensure
        \Statex $R$: Result pairs.
    \State $R\leftarrow \{\}$
    \State $S \leftarrow \lambda S^B + (1-\lambda) S^H$
    \While {$B^H \ne \O$}
        \State Record the highest scored pair as $T$
        \State Remove $T$ from $B^H$ and $B^B$, add it to $R$
        \For {$(b^H_i, b^B_i)\ \in\ (B^H,B^B)$}
            \State $overlap^H \leftarrow \mathrm{IoU}(T^H, b^H_i)$
            \State $overlap^B \leftarrow \mathrm{IoU}(T^B, b^B_i)$
            \If {$overlap^H>\Omega_H\ \mathbf{or}\ overlap^B > \Omega_B$}
                \State Remove $i$-th element from $B^H, S^H$
                \State Remove $i$-th element from $B^B, S^B$
            \EndIf
        \EndFor
        \EndWhile
            \State \Return{$R$}
    \end{algorithmic}
    \label{algm:JointNMS}
\end{algorithm}

\subsection{Feature Aggregation}
\label{sec:Aggregation}

Features of heads may significantly help discriminate instances from the crowd. In the meanwhile, semantic information from body will also benefit the head prediction by providing effective context. Therefore, features of heads and bodies are aggregated in Double Anchor R-CNN.

Aggregating features of heads and bodies have different ways. A simple solution is to directly combine the spatial feature maps or fully-connected~(FC) vectors together. In this work, we try both the two methods and choose the latter implementation to avoid the misalignments between head features and body features. Moreover, the classification task usually requires more global information and the localization task demands better spatial resolution. Therefore, we decouple the classification and localization tasks into two branches. The classification features of heads and bodies are extracted by the aggregated FC vectors. Regression tasks of heads and bodies are performed independently on individual feature maps, respectively.

\begin{table*}[!t]
    \centering
    \begin{center}
\begin{tabular}{c|c|cc|cc}
\hline
Method & Region Feature Extraction & MR-B & MR-H & $\Delta$MR-B & $\Delta$MR-H\\
\hline
Shao \etal \cite{CrowdHuman} & RoI Pooling & 55.94 & 52.06 & -0.58 & -1.72\\
Baseline (\cite{CrowdHuman} + RoI Align) & RoI Align & 55.36 & 50.34 & - & - \\
Baseline + Multi-task & RoI Align & 54.72 & - & +0.64 & - \\
Repulsion Loss~\cite{RepulsionLoss} & RoI Align & 54.64 & - & +0.72 & - \\
Soft-NMS~\cite{SoftNMS} & RoI Align & 60.05 & - & -7.30 & - \\
\hline
DA-RCNN & RoI Align & \textbf{52.30} & 49.98 & \textbf{+3.06} & +0.36\\
DA-RCNN + J-NMS & RoI Align & \textbf{51.79} & 49.68 & \textbf{+3.57} & +0.66\\
\hline
\end{tabular}
\end{center}
    \vspace{-0.3cm}
    \caption{Overall results on CrowdHuman \cite{CrowdHuman} val set. ``MR-B'' and ``MR-H'' stand for the log-average miss rate of visible body and head respectively. ``\cite{CrowdHuman}+RoI Align'' serves as our baseline. ``DA-RCNN'' in this table contains modules of Double Anchor RPN, Proposal Crossover and Feature Aggregation. ``Baseline + Multi-Task'' represents multi-class detection for head and body separately. On top of the DA-RCNN, Joint NMS (``J-NMS'') brings 0.49pp extra gains for human body and 0.3pp gains for head.}
    \vspace{-0.3cm}
    \label{table:overall}
\end{table*}

\subsection{Joint NMS}
Non-Maximum Suppression (NMS) is an essential step for removing duplicated predictions in detection frameworks. The performance of detectors is greatly affected by the NMS threshold, especially in crowded situations. Applying a higher threshold like 0.7 will increase false positives while a lower threshold like 0.3 may lead to a bad recall. 

In this work, Joint NMS is adopted to improve the robustness of the post-processing procedure of human detection in crowded scenes. One of the biggest problems of human detection in a crowd lies in a large number of false positive predictions with high confidences \cite{RepulsionLoss}. Therefore, we propose to suppress false positive predictions by taking both the head parts and body parts into consideration. To be specific, the confidences between the two parts will be weighted together, and boxes with lower confidence will be suppressed if either the head overlap or body overlap exceeds the threshold. The Joint NMS algorithm is formally described in Algorithm~\ref{algm:JointNMS}. 

The benefit of Joint NMS can be summarized in two aspects. First, joint score follows the idea of ensemble and is more reliable than a single score of human body. Second, the original NMS only takes one branch into consideration. False positives caused by the other branch are not suppressed. In contrast, Joint NMS suppresses false positives from both branches at the same time. As a result, the proposed Joint NMS is more robust to hyperparameters compared to the original NMS.

\section{Experimental Results}
We evaluate our approach on three human detection benchmarks: CrowdHuman~\cite{CrowdHuman}, COCOPersons~\cite{COCO} and CrowdPose~\cite{CrowdPose}.

\subsection{Datasets and Evaluation Metric}
\textbf{CrowdHuman Dataset.} The CrowdHuman dataset~\cite{CrowdHuman} is a human detection benchmark aimed at evaluating detectors in crowded scenarios. Different from other datasets for pedestrian detection such as Caltech~\cite{Caltech}, KITTI~\cite{KITTI} and CityPersons~\cite{CityPersons}, there are more crowded cases in CrowdHuman dataset and the average number of persons in an image is much larger. Three categories of bounding boxes annotations are provided: head bounding boxes, human visible-region bounding boxes and human full-body bounding boxes. Detecting visible-region is more difficult since the aspect ratios are more diverse than the full-body annotations. We benchmark the proposed method with the visible-region and head annotations. All the experiments are trained on the training set, and evaluated on the validation set.

\begin{figure}[!t]
\centering
\includegraphics[width=8cm]{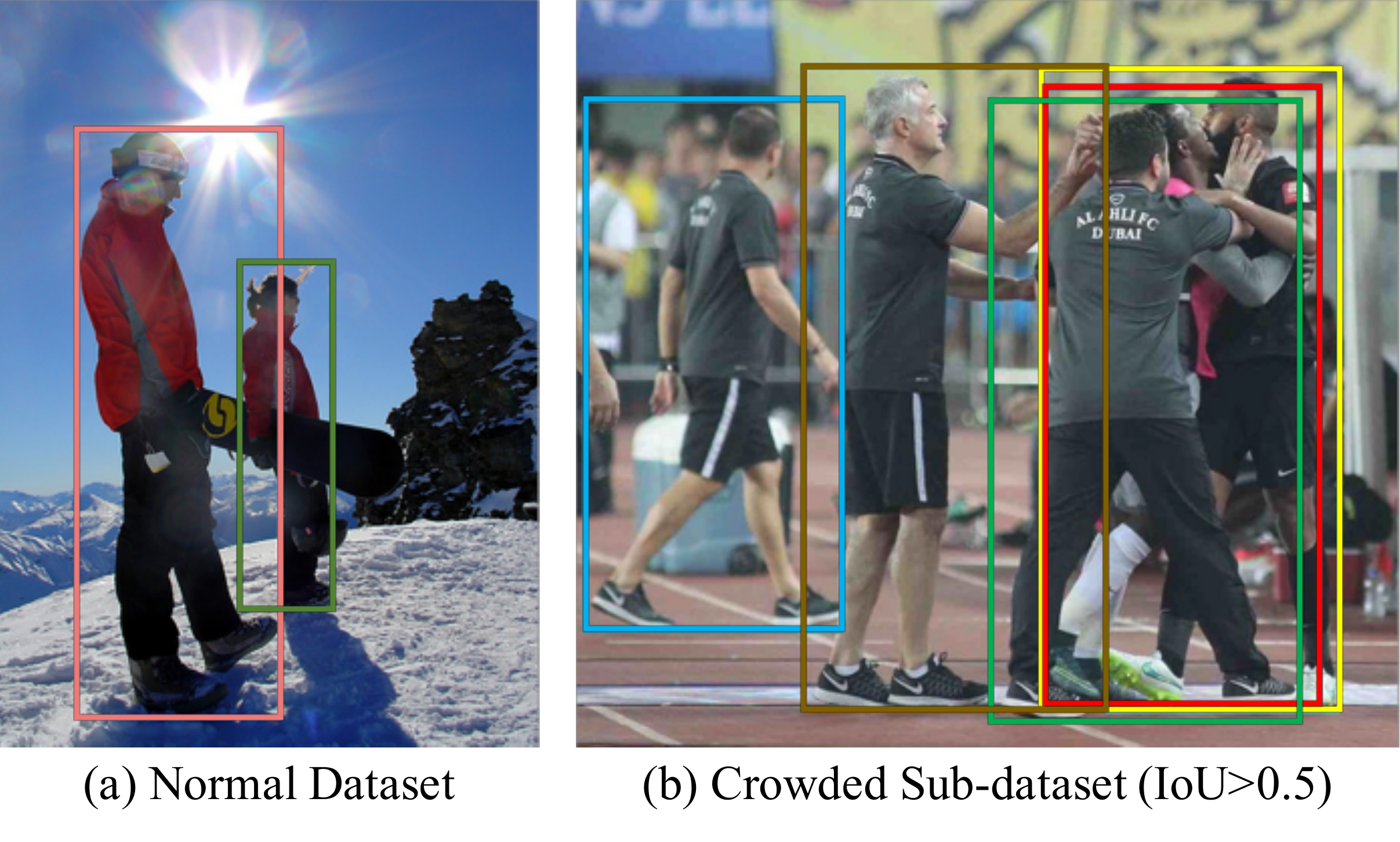}
\vspace{-0.5cm}
\caption{Visual comparisons between normal dataset and crowded sub-dataset of CrowdPose. The overlaps of humans are much larger in (b) which lead to increased difficulty in human detection.}
\vspace{-0.3cm}
\label{fig:crowd_subset}
\end{figure}

\textbf{COCOPersons and CrowdPose Dataset.} COCOPersons and CrowdPose are both benchmark datasets for human detection. COCOPersons is a subset of MSCOCO~\cite{COCO} from the images with ground-truth bounding boxes of ``person''. According to our statistics, there are 64115 images in the ``trainval minus minival'' dataset, and the ``minival'' has 2693 images for validation. CrowdPose~\cite{CrowdPose} is a recent dataset which extracts crowded images containing humans from MSCOCO~\cite{COCO}, MPII~\cite{MPII} and AI Challenger~\cite{AI-Challenger}. It should be noted that all the persons labeled in COCOPersons and CrowdPose are annotated like visible body and there aren't head bounding boxes annotations. To verify the effectiveness of our method, we annotate the head bounding boxes for persons in these two datasets. However, these two datasets are less crowded than CrowdHuman dataset, so we split out crowded sub-datasets with the images containing at least one pair of human boxes with an IoU greater than 0.5 from COCOPersons and CrowdPose, respectively. Visual comparisons between normal dataset and crowded sub-dataset of CrowdPose can be seen in Figure~\ref{fig:crowd_subset}.

\textbf{Evaluation Metric.} Standard log-average miss rate (MR)~\cite{Caltech} is chosen as a main metric in our experiments, which is the official metric of Caltech, CityPersons, and CrowdHuman dataset. The MR is computed in the false positive per image (FPPI) with a range of $[10^{-2}, 10^0]$ (${\rm MR}^{-2}$). Besides, AP$_{50}$ is also evaluated following the standard COCO evaluation metric.

\subsection{Implementation Details}
\label{sec:implementation}
We adopt FPN~\cite{FPN} with ResNet-50~\cite{ResNet} model pre-trained on ImageNet~\cite{ImageNet} dataset as our baseline. RoI Align~\cite{MaskRCNN} is adopted for better feature extraction. The head and visible body detection results for the baseline are obtained using two models trained for head and visible body separately. 
For all of CrowdHuman, COCOPersons and CrowdPose datasets, the anchor ratios for both human head and visible body detection are set to 1:2, 1:1, 2:1. Considering the various sizes of images in the dataset, the input image is re-scaled such that its shortest edge is 800 pixels, and the longest side is not beyond 1400 pixels. Synchronized SGD is adopted over 8 GPUs with a total of 16 images per minibatch and the initial learning rate is $0.02$. For CrowdHuman and CrowdPose dataset, we train 40 epochs in total and decrease the learning rate by 0.1 at epoch 20 and 30. As for COCOPersons dataset, we train 100k iterations in total and the learning rate is decreased by a factor of 10 after $60k$ and $80k$ iterations. 

\subsection{Detection Results on CrowdHuman}

\begin{figure*}[!t]
\centering
\includegraphics[width=17.2cm]{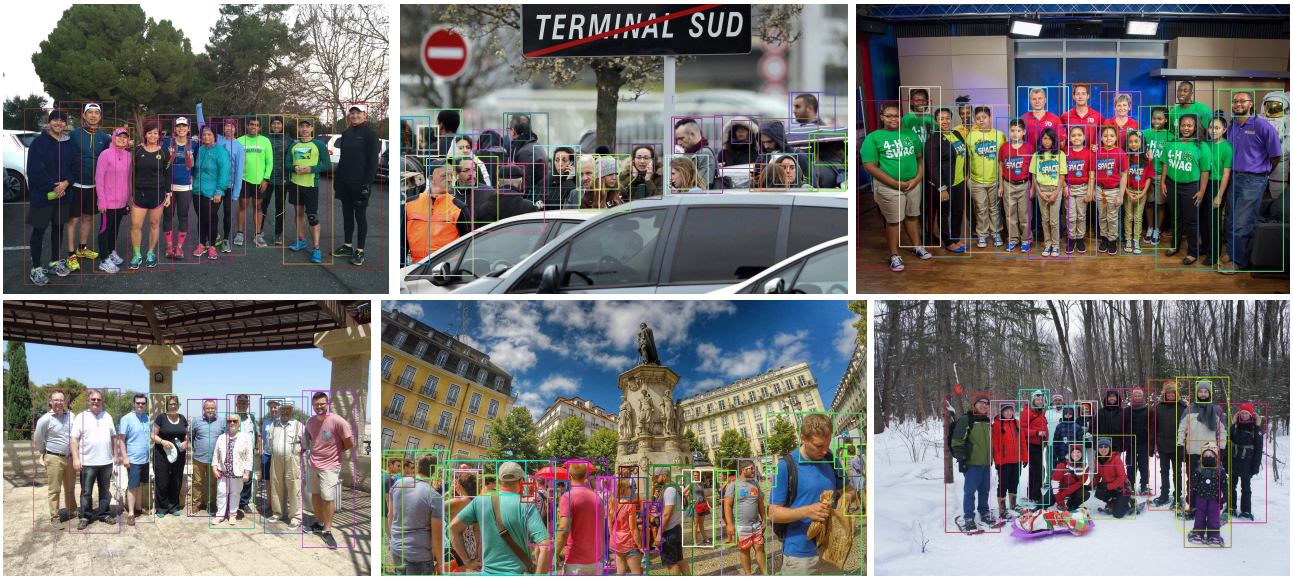}
\vspace{-0.4cm}
\caption{Qualitative results of Double Anchor R-CNN with Joint NMS on CrowdHuman (only predictions with confidences above 0.5 are drawn). Our method can effectively suppress false positives of humans, especially in crowded scenarios.}
\vspace{-0.3cm}
\label{fig:crowd_human_result}
\end{figure*}

\subsubsection{Overall Performance.}

\begin{table}[!t]
    \begin{center}
\begin{tabular}{l|c|c}
\hline
Method & MR-B & MR-H\\
\hline
Baseline (our implementation) & 55.36 & 50.34\\
\hline
DA-RPN, sample by head & 79.52 & 48.35\\
DA-RPN, sample by both-0.4 & 73.37 & 51.05\\
DA-RPN, sample by both-0.5 & 72.09 & 53.33\\
\hline
DA-RPN + crossover & \textbf{52.75} & 50.12\\
\hline
\end{tabular}
\end{center}
    \vspace{-0.4cm}
    \caption{Ablation studies on Proposal Crossover. ``DA-RPN'' stands for the proposed Double Anchor RPN. ``sample by head'' indicates that sampling positive proposals only uses head parts only. ``sample by both'' means the positive pair are selected using both head and person proposals. ``+crossover'' represents Proposal Crossover module is adopted.}
    \vspace{-0.2cm}
    \label{table:DoubleAnchorRPN}
\end{table}

\begin{table}[!t]
    \centering
    \begin{center}
\begin{tabular}{c|c|c|c|c}
\hline
Method & SP-Agg & FC-Agg & MR-B & MR-H\\
\hline
\multirow{3}{*}{DA-RCNN} & & & 52.75 & 50.12\\
 & \checkmark & & 52.90 & 52.70\\
 & & \checkmark & 52.30 & 49.98\\
\hline
\end{tabular}
\end{center}
    \vspace{-0.3cm}
    \caption{Ablation studies on Feature Aggregation. Spatial aggregation (``SP-Agg'') and FC vector aggregation (``FC-Agg'') are adopted for comparison.}
    \vspace{-0.3cm}
    \label{table:Aggregation}
\end{table}

The detection results on CrowdHuman are shown in Table~\ref{table:overall}. FPN and FPN with RoI Align are tested with original NMS on the head and visible body separately.
For the performance of body detection represented by ``MR-B'', DA-RCNN makes an improvement of 3.06pp compared to the baseline result. To further demonstrate that the performance improvement gains mainly from our method rather than collecting more annotations for the head boxes, we compare our method DA-RCNN with the multi-task learning, which detects heads and bodies as a multi-category task. DA-RCNN makes an improvement of 2.42pp compared to the multi-task learning. Moreover, Joint NMS can bring extra gains of 0.49pp for the human body detection based on the DA-RCNN, while the results of Soft-NMS is not optimistic. We argue that Soft-NMS maintains lots of long-tail detection results for improving recall at the expense of bringing more false positives, which leads to negative impact on human detection especially for the metric of MR. It is worth noting that the DA-RCNN with Joint NMS can surpass state-of-the-art method using Repulsion Loss on CrowdHuman dataset for human body detection, which indicates the effectiveness of our method to detect the human in crowded scenes. Besides, the performance of head detection is improved by 0.36pp, benefiting from the context information provided by human body. Example results from our method are visualized in Figure~\ref{fig:crowd_human_result}.

\begin{table}[!t]

\begin{center}
\begin{tabular}{c|c|c|c}
\hline
NMS Type & $\lambda$ & MR-B & $\Delta$\\
\hline
Original NMS & 1.0 & 52.30 & - \\
\hline
Joint NMS & 0.9 & 51.83 & +0.47\\
Joint NMS & 0.8 & \textbf{51.79} & \textbf{+0.51}\\
Joint NMS & 0.7 & 51.95 & +0.35\\
Joint NMS & 0.6 & 52.06 & +0.24\\
Joint NMS & 0.5 & 52.16 & +0.14\\
\hline
\end{tabular}
\end{center}
    \vspace{-0.4cm}
    \caption{Ablation studies on Joint NMS for DA-RCNN.}
    \vspace{-0.2cm}
    \label{table:JointNMS}
\end{table}

\begin{table}[!t]
    \begin{center}
\begin{tabular}{c|cccc}
\hline
Recall & 0.2 & 0.4 & 0.6 & 0.8 \\
\hline
Baseline & 0.0120 & 0.0933 & 0.5260 & 3.2976 \\
Ours & \textbf{0.0098} & \textbf{0.0664} & \textbf{0.3654} & \textbf{2.1121} \\
\hline
\end{tabular}
\end{center}
    \vspace{-0.4cm}
    \caption{False positive per image (FPPI) under various recall rates. Our method is effective in reducing false positives compared to the baseline.}
    \vspace{-0.3cm}
    \label{table:recall_fppi}
\end{table}

\subsubsection{Ablation Study on Proposal Crossover.}
\label{sec:Selection}

We evaluate different proposal selecting strategies for Double Anchor R-CNN. The results are illustrated in Table \ref{table:DoubleAnchorRPN}. 
The naive implementation samples positive proposals according to the head parts only (termed as ``DA-RPN, sample by head''). The method brings an improvement of $~2$pp on MR for heads, which indicates that constructing the relationship between head and corresponding body is beneficial to head detection. However, as discussed in Section~\ref{sec:Crossover}, sampling proposals by head parts will sacrifice the performance of human body detection since the body proposals are noisy.

Then the sampling strategy switches to an updated version which takes both head and body proposals into account. The method is represented as ``sample by both-x'' in Table~\ref{table:DoubleAnchorRPN} and ``x'' stands for the positive overlap threshold for person boxes. Obviously, the MR of body (MR-B) is significantly improved compared to the naive sampling strategy. Note that with the increasing overlap threshold, the result for visible body detection is better while the result is worse for head detection. This indicates the trade-off between the number of noisy samples for visible body and the decrease in the number of positive proposals. However, the human detection result is still much worse than the baseline results since the reduction in the number of qualified proposals is very harmful to the detection performance.

Finally, we adopt a proposal crossover module to improve the quantity and quality of paired proposals. Shown as ``+crossover'' in Table~\ref{table:DoubleAnchorRPN}, the proposal crossover module brings a significant improvement of $19.34$pp for the result of body detection. 
The improvement is benefited from the increasing number of qualified pairs of proposals provided by the proposal crossover module. 
To prove the assumption, we calculate the number of qualified pairs of proposals in training. There are only $\sim$40 positive pairs per image on average if proposals are sampled by requiring a threshold of 0.5 IoU for both body and head parts. In contrast, the average number of positive proposal pairs after the crossover strategy increases to 97 per image.
It proved that more qualified proposals are beneficial to detection performance.

\subsubsection{Ablation Study on Feature Aggregation.}


As discussed in Section~\ref{sec:Aggregation}, we adopt FC vectors aggregation module in our work. The results are illustrated in Table~\ref{table:Aggregation}. Compared with the baseline framework without feature aggregation module, fusing FC vectors leads to a gain of $0.45$pp on MR-B and also an improvement of $0.14$pp on MR-H. The results prove the effectiveness of feature aggregation. Besides, compared to aggregating features with FC vectors, fusing spatial feature maps leads to a drop of $2.58$pp on MR-H because of the misalignments of head and body features. 

\begin{table}[!t]


\begin{center}
\begin{tabular}{c|c|c|c}
\hline
Dataset & Method & MR & AP$_{50}$ \\
\hline
\multirow{2}*{COCOPersons-All} & Baseline & 39.36 & 84.23 \\
& Ours & 37.97 & 84.54 \\
\hline
\multirow{2}*{COCOPersons-Crowd} & Baseline & 58.83 & 79.25 \\
& Ours & \textbf{55.01} & \textbf{80.53} \\
\hline
\multirow{2}*{CrowdPose-All} & Baseline & 40.71 & 84.03 \\
& Ours & 39.12 & 84.45 \\
\hline
\multirow{2}*{CrowdPose-Crowd} & Baseline & 44.26 & 80.78 \\
& Ours & \textbf{40.02} & \textbf{84.56} \\
\hline
\end{tabular}
\end{center}
\vspace{-0.4cm}
\caption{Experiments on COCOPersons and CrowdPose. ``All'' stands for the whole validation dataset and ``Crowd'' stands for crowded sub-dataset.}
\vspace{-0.4cm}
\label{table:COCOPersons_CrowdPose}
\end{table}


\subsubsection{Ablation Study on Joint NMS.}
To prove the validity of Joint NMS, we compare it with original NMS on the human body detection task in Table~\ref{table:JointNMS}. Threshold of original NMS is set to 0.5 for simplicity.
As for the Joint NMS, the weighting factor $\lambda$ is a hyper-parameter for balancing the head scores and visible body scores.
Different values of $\lambda$ are evaluated and the result of visible body detection becomes better as the weight of body score increases. We are also able to find that the result is not sensitive to this factor. Moreover, to validate the effectiveness of suppressing false positives, we compare the results under ``FPPI over recall'' in Table~\ref{table:recall_fppi}. It is obvious that the proposed method is helpful to reduce false positive effectively under almost all recall settings. 


\subsection{Results on COCOPersons and CrowdPose}

To investigate the generalization capacity of the proposed methods, experimental results on COCOPersons and CrowdPose are reported in Table~\ref{table:COCOPersons_CrowdPose}.
The proposed Double Anchor R-CNN with Joint NMS is able to improve MR by 1.39pp and 1.59pp on the whole validation datasets of COCOPersons and CrowdPose, respectively. Compared with CrowdHuman, the COCOPersons and CrowdPose dataset are less crowded. As a result, we split out a crowded sub-dataset consisting of images containing at least one pair of human boxes with an IoU greater than 0.5. For the crowded sub-dataset, our method can achieve a huge boost of 3.82pp on MR and 1.28 point on AP$_{50}$ for COCOPersons, and a healthy 4.24pp MR gap and 3.78 point AP$_{50}$ gap for CrowdPose. The results demonstrate that the proposed framework is also suitable for regular challenging human detection dataset and is more effective on crowded scenarios.


\section{Conclusion}
We propose Double Anchor R-CNN for human detection in crowded scenes. 
The framework is intuitive and effective for handling crowd occlusion problem by naturally coupling the head and body for each person. Through a variety of experiments on challenging human detection datasets, Double Anchor R-CNN is demonstrated to be capable of improving performance and producing a state-of-the-art performance. 
Our approach is also extensive and can be easily generalized to detect other parts, for example, detecting the head, face and body of each person with triple anchor R-CNN. We hope the proposed method provides insights into future works on human detection and human-object interactions.

\bibliography{Bibliography-File}

\begin{thebibliography}{}

\bibitem[\protect\citeauthoryear{Alahi, Ramanathan, and
  Li}{2014}]{Socially-Aware}
Alahi, A.; Ramanathan, V.; and Li, F.~F.
\newblock 2014.
\newblock Socially-aware large-scale crowd forecasting.
\newblock In {\em CVPR}.

\bibitem[\protect\citeauthoryear{Andriluka \bgroup et al\mbox.\egroup
  }{2014}]{MPII}
Andriluka, M.; Pishchulin, L.; Gehler, P.; and Schiele, B.
\newblock 2014.
\newblock 2d human pose estimation: New benchmark and state of the art
  analysis.
\newblock In {\em CVPR}.

\bibitem[\protect\citeauthoryear{Bodla \bgroup et al\mbox.\egroup
  }{2017}]{SoftNMS}
Bodla, N.; Singh, B.; Chellappa, R.; and Davis, L.~S.
\newblock 2017.
\newblock Soft-nms -- improving object detection with one line of code.
\newblock In {\em The IEEE International Conference on Computer Vision (ICCV)}.

\bibitem[\protect\citeauthoryear{Cai and Vasconcelos}{2018}]{CascadeRCNN}
Cai, Z., and Vasconcelos, N.
\newblock 2018.
\newblock Cascade r-cnn: Delving into high quality object detection.
\newblock In {\em CVPR}.

\bibitem[\protect\citeauthoryear{Cai \bgroup et al\mbox.\egroup }{2016}]{MSCNN}
Cai, Z.; Fan, Q.; Feris, R.; and Vasconcelos, N.
\newblock 2016.
\newblock A unified multi-scale deep convolutional neural network for fast
  object detection.
\newblock In {\em ECCV}.

\bibitem[\protect\citeauthoryear{Deng \bgroup et al\mbox.\egroup
  }{2009}]{ImageNet}
Deng, J.; Dong, W.; Socher, R.; Li, L.-J.; Li, K.; and Fei-Fei, L.
\newblock 2009.
\newblock Imagenet: A large-scale hierarchical image database.
\newblock In {\em CVPR}.

\bibitem[\protect\citeauthoryear{Dollar \bgroup et al\mbox.\egroup
  }{2012}]{Caltech}
Dollar, P.; Wojek, C.; Schiele, B.; and Perona, P.
\newblock 2012.
\newblock Pedestrian detection: An evaluation of the state of the art.
\newblock {\em T-PAMI} 34(4):743--761.

\bibitem[\protect\citeauthoryear{Duan, Ai, and Lao}{2010}]{Duan2010A}
Duan, G.; Ai, H.; and Lao, S.
\newblock 2010.
\newblock A structural filter approach to human detection.
\newblock In {\em ECCV}.

\bibitem[\protect\citeauthoryear{Enzweiler \bgroup et al\mbox.\egroup
  }{2010}]{Enzweiler2010Multi}
Enzweiler, M.; Eigenstetter, A.; Schiele, B.; and Gavrila, D.~M.
\newblock 2010.
\newblock Multi-cue pedestrian classification with partial occlusion handling.
\newblock In {\em CVPR}.

\bibitem[\protect\citeauthoryear{Everingham \bgroup et al\mbox.\egroup
  }{2010}]{PASCAL}
Everingham, M.; Van~Gool, L.; Williams, C. K.~I.; Winn, J.; and Zisserman, A.
\newblock 2010.
\newblock The pascal visual object classes (voc) challenge.
\newblock {\em IJCV} 88(2):303--338.

\bibitem[\protect\citeauthoryear{Geiger, Lenz, and Urtasun}{2012}]{KITTI}
Geiger, A.; Lenz, P.; and Urtasun, R.
\newblock 2012.
\newblock Are we ready for autonomous driving? the kitti vision benchmark
  suite.
\newblock In {\em CVPR}.

\bibitem[\protect\citeauthoryear{He \bgroup et al\mbox.\egroup }{2016}]{ResNet}
He, K.; Zhang, X.; Ren, S.; and Sun, J.
\newblock 2016.
\newblock Deep residual learning for image recognition.
\newblock In {\em CVPR}.

\bibitem[\protect\citeauthoryear{He \bgroup et al\mbox.\egroup
  }{2017}]{MaskRCNN}
He, K.; Gkioxari, G.; Dollar, P.; and Girshick, R.
\newblock 2017.
\newblock Mask r-cnn.
\newblock In {\em ICCV}.

\bibitem[\protect\citeauthoryear{Li \bgroup et al\mbox.\egroup
  }{2017}]{SAF-RCNN}
Li, J.; Liang, X.; Shen, S.; Xu, T.; Feng, J.; and Yan, S.
\newblock 2017.
\newblock Scale-aware fast r-cnn for pedestrian detection.
\newblock {\em T-MM} 20(4):985--996.

\bibitem[\protect\citeauthoryear{Li \bgroup et al\mbox.\egroup
  }{2019}]{CrowdPose}
Li, J.; Wang, C.; Zhu, H.; Mao, Y.; Fang, H.-S.; and Lu, C.
\newblock 2019.
\newblock Crowdpose: Efficient crowded scenes pose estimation and a new
  benchmark.
\newblock In {\em CVPR}.

\bibitem[\protect\citeauthoryear{Lin \bgroup et al\mbox.\egroup }{2014}]{COCO}
Lin, T.-Y.; Maire, M.; Belongie, S.; Hays, J.; Perona, P.; Ramanan, D.;
  Dollár, P.; and Zitnick, C.~L.
\newblock 2014.
\newblock Microsoft coco: Common objects in context.
\newblock In {\em ECCV}.

\bibitem[\protect\citeauthoryear{Lin \bgroup et al\mbox.\egroup }{2017}]{FPN}
Lin, T.-Y.; Dollar, P.; Girshick, R.; He, K.; Hariharan, B.; and Belongie, S.
\newblock 2017.
\newblock Feature pyramid networks for object detection.
\newblock In {\em CVPR}.

\bibitem[\protect\citeauthoryear{Lin \bgroup et al\mbox.\egroup
  }{2018}]{Graininess-Aware}
Lin, C.; Lu, J.; Wang, G.; and Zhou, J.
\newblock 2018.
\newblock Graininess-aware deep feature learning for pedestrian detection.
\newblock In {\em ECCV}.

\bibitem[\protect\citeauthoryear{Liu \bgroup et al\mbox.\egroup }{2016}]{SSD}
Liu, W.; Anguelov, D.; Erhan, D.; Szegedy, C.; Reed, S.; Fu, C.-Y.; and Berg,
  A.~C.
\newblock 2016.
\newblock {SSD}: Single shot multibox detector.
\newblock In {\em ECCV}.

\bibitem[\protect\citeauthoryear{Liu, Huang, and Wang}{2019}]{AdaptiveNMS}
Liu, S.; Huang, D.; and Wang, Y.
\newblock 2019.
\newblock Adaptive nms: Refining pedestrian detection in a crowd.
\newblock In {\em CVPR}.

\bibitem[\protect\citeauthoryear{Mathias \bgroup et al\mbox.\egroup
  }{2014}]{Mathias2014Handling}
Mathias, M.; Benenson, R.; Timofte, R.; and Gool, L.~V.
\newblock 2014.
\newblock Handling occlusions with franken-classifiers.
\newblock In {\em ICCV}.

\bibitem[\protect\citeauthoryear{Noh \bgroup et al\mbox.\egroup
  }{2018}]{ImprovingOcclusion_PedDet}
Noh, J.; Lee, S.; Kim, B.; and Kim, G.
\newblock 2018.
\newblock Improving occlusion and hard negative handling for single-stage
  pedestrian detectors.
\newblock In {\em CVPR}.

\bibitem[\protect\citeauthoryear{Ouyang and Wang}{2012}]{Ouyang2012A}
Ouyang, W., and Wang, X.
\newblock 2012.
\newblock A discriminative deep model for pedestrian detection with occlusion
  handling.
\newblock In {\em CVPR}.

\bibitem[\protect\citeauthoryear{Ouyang and Wang}{2014}]{Ouyang2014Joint}
Ouyang, W., and Wang, X.
\newblock 2014.
\newblock Joint deep learning for pedestrian detection.
\newblock In {\em ICCV}.

\bibitem[\protect\citeauthoryear{Redmon and Farhadi}{2017}]{YOLO9000}
Redmon, J., and Farhadi, A.
\newblock 2017.
\newblock Yolo9000: Better, faster, stronger.
\newblock In {\em CVPR}.

\bibitem[\protect\citeauthoryear{Redmon \bgroup et al\mbox.\egroup
  }{2016}]{YOLO}
Redmon, J.; Divvala, S.; Girshick, R.; and Farhadi, A.
\newblock 2016.
\newblock You only look once: Unified, real-time object detection.
\newblock In {\em CVPR}.

\bibitem[\protect\citeauthoryear{Ren \bgroup et al\mbox.\egroup
  }{2015}]{FasterRCNN}
Ren, S.; He, K.; Girshick, R.; and Sun, J.
\newblock 2015.
\newblock Faster {R-CNN}: Towards real-time object detection with region
  proposal networks.
\newblock In {\em NIPS}.

\bibitem[\protect\citeauthoryear{Shao \bgroup et al\mbox.\egroup
  }{2018}]{CrowdHuman}
Shao, S.; Zhao, Z.; Li, B.; Xiao, T.; Yu, G.; Zhang, X.; and Sun, J.
\newblock 2018.
\newblock Crowdhuman: A benchmark for detecting human in a crowd.
\newblock {\em arXiv:1805.00123}.

\bibitem[\protect\citeauthoryear{Tian \bgroup et al\mbox.\egroup
  }{2016}]{Tian2016Deep}
Tian, Y.; Luo, P.; Wang, X.; and Tang, X.
\newblock 2016.
\newblock Deep learning strong parts for pedestrian detection.
\newblock In {\em ICCV}.

\bibitem[\protect\citeauthoryear{Wang \bgroup et al\mbox.\egroup
  }{2018}]{RepulsionLoss}
Wang, X.; Xiao, T.; Jiang, Y.; Shao, S.; Sun, J.; and Shen, C.
\newblock 2018.
\newblock Repulsion loss: Detecting pedestrians in a crowd.
\newblock In {\em CVPR}.

\bibitem[\protect\citeauthoryear{Wang}{2012}]{Wang2012A}
Wang, X.
\newblock 2012.
\newblock A discriminative deep model for pedestrian detection with occlusion
  handling.
\newblock In {\em CVPR}.

\bibitem[\protect\citeauthoryear{Wu \bgroup et al\mbox.\egroup
  }{2017}]{AI-Challenger}
Wu, J.; Zheng, H.; Zhao, B.; Li, Y.; Yan, B.; Liang, R.; Wang, W.; Zhou, S.;
  Lin, G.; Fu, Y.; Wang, Y.; and Wang, Y.
\newblock 2017.
\newblock {AI} challenger : {A} large-scale dataset for going deeper in image
  understanding.
\newblock {\em arXiv:1711.06475}.

\bibitem[\protect\citeauthoryear{Zhang, Benenson, and
  Schiele}{2017}]{CityPersons}
Zhang, S.; Benenson, R.; and Schiele, B.
\newblock 2017.
\newblock Citypersons: A diverse dataset for pedestrian detection.
\newblock In {\em CVPR}.

\bibitem[\protect\citeauthoryear{Zhang \bgroup et al\mbox.\egroup
  }{2016}]{HowFar_PedDet}
Zhang, S.; Benenson, R.; Omran, M.; Hosang, J.; and Schiele, B.
\newblock 2016.
\newblock How far are we from solving pedestrian detection?
\newblock In {\em CVPR}.

\bibitem[\protect\citeauthoryear{Zhang \bgroup et al\mbox.\egroup
  }{2018}]{Occlusion-aware}
Zhang, S.; Wen, L.; Bian, X.; Lei, Z.; and Li, S.~Z.
\newblock 2018.
\newblock Occlusion-aware r-cnn: Detecting pedestrians in a crowd.
\newblock In {\em ECCV}.

\bibitem[\protect\citeauthoryear{Zhang, Yang, and
  Schiele}{2018}]{GuidedAttention}
Zhang, S.; Yang, J.; and Schiele, B.
\newblock 2018.
\newblock Occluded pedestrian detection through guided attention in cnns.
\newblock In {\em CVPR}.

\bibitem[\protect\citeauthoryear{Zhou and Yuan}{2016}]{Zhou2016Learning}
Zhou, C., and Yuan, J.
\newblock 2016.
\newblock Learning to integrate occlusion-specific detectors for heavily
  occluded pedestrian detection.
\newblock In {\em ACCV}.

\bibitem[\protect\citeauthoryear{Zhou and Yuan}{2017}]{Zhou2017Multi}
Zhou, C., and Yuan, J.
\newblock 2017.
\newblock Multi-label learning of part detectors for heavily occluded
  pedestrian detection.
\newblock In {\em ICCV}.

\bibitem[\protect\citeauthoryear{Zhou and Yuan}{2018}]{Bi-box}
Zhou, C., and Yuan, J.
\newblock 2018.
\newblock Bi-box regression for pedestrian detection and occlusion estimation.
\newblock In {\em ECCV}.

\end{thebibliography}
\bibliographystyle{aaai}
\end{document}